\documentclass{article}

\usepackage{arxiv}
\usepackage[utf8]{inputenc} 
\usepackage[T1]{fontenc}
\usepackage{xspace}
\usepackage{amsfonts}
\usepackage{amsmath}
\usepackage{cancel}
\usepackage{xstring} 

\usepackage{graphicx}
\usepackage{verbatim}
\usepackage{graphicx}
\usepackage{url}            
\usepackage{lipsum}		
\usepackage{graphicx}
\usepackage{doi}
\usepackage{multicol}
\usepackage{multirow} 
\usepackage{caption}
\usepackage{hyperref}
\usepackage{authblk}
\newtheorem{lemma}{Lemma}
\newtheorem{proof}{Proof}

\title{Multi-Response Preference Optimization with Augmented Ranking Dataset}

\author[1,4]{Hansle Gwon}
\author[1,5]{Imjin Ahn}
\author[2]{Young-Hak Kim}
\author[4]{Sanghyun Park}
\author[3]{Tae Joon Jun}

\affil[1]{INMED DATA, 88, Olympicro 43gil, Songpagu, 05505, Seoul, Republic of Korea}
\affil[2]{Division of Cardiology, Department of Information Medicine, Asan Medical Center, University of Ulsan College of Medicine,  88, Olympicro 43gil, Songpagu, 05505, Seoul, Republic of Korea}
\affil[3]{Big Data Research Center, Asan Institute for Life Sciences, Asan Medical Center, 88, Olympicro 43gil, Songpagu, 05505, Seoul, Republic of Korea}
\affil[4]{Department of Computer Science, Yonsei University, Seoul, Republic of Korea}
\affil[5]{Department of Artificial Intelligence, Yonsei University, Seoul, Republic of Korea}

\hypersetup{
pdftitle={Multi-Response Preference Optimization with Augmented Ranking Dataset},
pdfsubject={Machine Learning},
pdfkeywords={Large Language Models, Preference Optimization , Natural Language Process},
}

\begin{document}
\maketitle

\begin{abstract}
Recent advancements in Large Language Models (LLMs) have been remarkable, with new models consistently surpassing their predecessors. These advancements are underpinned by extensive research on various training mechanisms. Among these, Preference Optimization has played a significant role in improving the performance of LLMs by incorporating human preferences into the training process. However, constructing preference optimization datasets is challenging and the optimization process is highly sensitive to the dataset quality. In this study, we propose a novel approach to augment Preference Optimization datasets. Additionally, we introduce a Multi-response-based Preference Optimization training method that enables the simultaneous learning of multiple responses.

\end{abstract}

\section{Introduction}
Transformers\cite{vaswani2017attention} have revolutionized natural language processing (NLP) since their introduction, setting numerous state-of-the-art (SOTA). Using the attention mechanism, this model demonstrated superior performance compared to existing models in most NLP tasks. Furthermore, Transformers enabled large-scale data training through parallelization. Thanks to their parallelizable architecture, it became possible to pre-train on massive amounts of unlabeled data. These pre-trained models, when fine-tuned with labeled data, brought about significant advancements once again.

The success of Transformers led to the emergence of Large Language Models (LLM)\cite{brown2020language,touvron2023llama}. LLMs are developed by training Transformer models with billions of parameters on vast amounts of data. These massive models acquire knowledge across various domains and can perform a wide range of tasks using a single model. Unlike previous generations of Transformers, which were fine-tuned for specific tasks, LLMs have been optimized to follow human instructions through a process known as instruction fine-tuning. Instruction fine-tuning\cite{ouyang2022traininglanguagemodelsfollow,alpaca} is one of the most prominent methods for fine-tuning LLMs. In this approach, pre-trained LLMs are further trained on datasets comprising instructions and their corresponding responses. As a result, fine-tuned LLMs for instruction become significantly better at executing human directives.

Although instruction fine-tuning significantly enhanced LLMs' ability to follow instructions, some challenges remained. Instruction fine-tuned models strictly adhere to directives without discerning their correctness. Consequently, such models may produce harmful outputs, including misinformation present in indiscriminately collected training data. Human preference optimization (preference alignment) effectively mitigates this issue. By suppressing undesirable responses and promoting preferred ones, human preference optimization trains LLMs to generate responses aligned with human preferences while avoiding harmful outputs.
Reinforcement Learning from Human Feedback(RLHF)\cite{ouyang2022training} is the most well-known method for incorporating human preferences. It optimizes models to maximize rewards determined by a reward model through reinforcement learning. RLHF has proven effective not only in suppressing harmful outputs but also in improving model performance. However, RLHF involves a complex training process, as it requires the additional training of a reward model.
Direct Preference Optimization(DPO)\cite{rafailov2024direct} addresses RLHF's complexity by directly optimizing models using preference datasets. DPO simplifies the preference optimization process while achieving excellent performance, making it a notable alternative to RLHF. Nevertheless, both RLHF and DPO face challenges in data acquisition. These methods rely on human-labeled preference datasets, which are costly and time-intensive to construct, making large-scale dataset creation difficult.

In this study, we propose a method for augmenting preference datasets to construct large-scale preference datasets using commercial models such as GPT-4\cite{openai2024gpt4technicalreport}. Our approach generates data similar in format to an initial preference dataset and performs preference labeling through the model. This process enables the large-scale augmentation of small datasets without human intervention.
Additionally, unlike existing DPO methods that rely on two responses (chosen and rejected), we propose a training method that simultaneously incorporates multiple responses. By utilizing multiple responses, our approach captures human preferences in greater detail, leading to more robust learning.
Through experiments, we validate the effectiveness of the two proposed methods—data augmentation and multi-response preference learning.

\section{Preliminary}
\subsection{RLHF}
In RLHF, pretrained LLMs undergo supervised fine-tuning (SFT)\cite{ouyang2022training} using high-quality datasets. Once fine-tuning is completed, the SFT model, $\pi^{SFT}$, learns human preferences through reinforcement learning. For reinforcement learning, it is necessary to train a reward model $r^{*}$ that provides preference feedback.
The Bradley-Terry (BT)\cite{bradley1952rank} model represents the probability distribution for the preference of response $y_{1}$ over $y_{2}$ for a given prompt $x$ as follows:

\begin{equation}
    p^{*}(y_{1}>y_{2} | x) = \frac{\exp(r^{*}(x,y_{1}))}{\exp(r^{*}(x,y_{1})) + \exp(r^{*}(x,y_{2}))}
\label{equation:RLHF_BT}
\end{equation}

$r^{*}$ can be replaced with a parameterized model $r_{\phi}$, and based on equation \ref{equation:RLHF_BT}, the following loss function can be derived.

\begin{equation}
    \mathcal{L}_{R}(r_{\phi}, \mathcal{D}) =  \mathbb{E}_{(x, y_{w}, y_{l}) \sim \mathcal{D}}[\log\sigma(r_{\phi}(x, y_{w}) - r_{\phi}(x, y_{l}))]
\label{equation:RLHF_reward}
\end{equation}

In the dataset $\mathcal{D} = \{x^{i}_{}, y^{i}_{w},y^{i}_{l}\}^{N}_{i=1}$, $y_{w}$ represents the preferred response, and $y_{l}$ represents the dispreferred response. The reward model is trained according to equation \ref{equation:RLHF_reward}, and the trained reward model provides feedback to the model during the RL stage.

\begin{equation}
    \max_{\pi_{\theta}}\mathbb{E}_{(x \sim \mathcal{D}, y \sim \pi_{\theta}(y|x))}[r_{\phi}(x,y)] - \beta \mathbb{D}_{KL}[\pi_{\theta}(y|x)||\pi_{ref}(y|x)]
\label{equation:RLHF_tuning}
\end{equation}

\subsection{Direct Preference Optimization}
DPO, inspired by the RLHF approach to learning human preferences through reinforcement learning, differs by directly learning preferences from the data without the intervention of rewards. According to prior studies \cite{go2023aligning,korbak2022reinforcement,peng2019advantage,peters2007reinforcement}, the optimal solution for equation \ref{equation:RLHF_tuning} can be expressed as follows:
\begin{equation}
    \pi_{r}(y|x) = \frac{1}{Z(x)}\pi_{ref}(y|x)\exp(\frac{1}{\beta}r(x,y))
\end{equation}
From this point, we can derive an equation for r(x,y).
\begin{equation}
    r(x,y) = \beta \log\frac{\pi_{r}(y|x)}{\pi_{ref}(y|x)} + \beta \log(Z(x))
\label{equation:new_reward}
\end{equation}

According to equation \ref{equation:new_reward}, equation \ref{equation:RLHF_BT} is defined as follows for the optimal policy $\pi^{*}$.

\begin{equation}
    p^{*}(y_{1} > y_{2} | x) = \frac{1}{1+\exp(\beta \log \frac{\pi^{*}(y_{2}|x)}{\pi_{ref}(y_{2}|x)} - \beta \log \frac{\pi^{*}(y_{1}|x)}{\pi_{ref}(y_{1}|x)})} 
\label{equation:DPO_probability}
\end{equation}

Based on equation \ref{equation:DPO_probability}, the objective function of DPO is defined as follows.

\begin{equation}
    \max_{\pi_{\theta}}\mathbb{E}_{(x,y_{w},y_{l}) \sim \mathcal{D}}[\log\sigma(\beta \frac{\pi_{\theta}(y_{w}|x)}{\pi_{ref}(y_{w}|x)} - \beta \frac{\pi_{\theta}(y_{l}|x)}{\pi_{ref}(y_{l}|x)})] 
\label{equation:DPO_tuning}
\end{equation}

\section{Related Works}
\subsection{Model rewarding preference optimization}

Human preference optimization significantly enhances model performance but requires datasets with preferences labeled by humans. The construction of such datasets is both time-consuming and costly, representing a major limitation of preference optimization. RLAIF (Reinforcement Learning from AI Feedback)\cite{lee2023rlaif} addresses this challenge by employing an LLM labeler to learn preferences. Preferences were evaluated using PaLM, and reinforcement learning was conducted with LLM-labeled data, achieving performance comparable to RLHF.

In the study by Yuan, Weizhe, et al.\cite{yuan2024self}, the model autonomously generates and evaluates data. During each iteration, the model creates new responses, evaluates them, and uses the resulting preference pairs to perform preference optimization. This iterative process allows the model to learn from increasingly improved data at each step, leading to continual enhancement.

The research above, along with the LLMs-as-a-judge approach, is heavily influenced by the performance of the LLMs. However, LLMs are generally not optimized to perform the role of a judge. Additionally, since the output of the judge LLM is text rather than a scalar value, it can be subjective. 

\subsection{Multi-DPO}
In the study by Song et al., Preference Ranking Optimization (PRO)\cite{song2024preference} introduces a method to optimize preferences using ranked datasets. PRO extends the DPO loss to accommodate datasets with ranking information. The loss in PRO is designed to maximize the probabilities for rank comparisons, including from the 1st to the \textit{n}-th rank, the 2nd to the \textit{n}-th rank, and so on, up to the \textit{(n-1)}-th to \textit{n}-th ranks. Additionally, the top-ranked samples are utilized for Supervised Fine-Tuning (SFT).


The study by Bai, Zhuoxi, et al., titled Direct Multi-Preference Optimization (DMPO)\cite{bai2024aligning}, proposes a preference optimization method tailored for recommendations. DMPO trains the model using one positive sample and multiple negative samples. The objective is to maximize the probability of the positive sample while minimizing the probabilities of the negative samples. Additionally, the positive samples are also used for supervised fine-tuning (SFT).

The studies above performed computation for all the comparisons set in the experiments. This approach can become inefficient as the amount of data increases.

\section{Methods}
In this study, we propose a method that augments datasets for preference optimization and conducts optimization using an improved approach. Consequently, our method is divided into two stages: data augmentation and training.
\subsection{Preference Dataset augmentation}

\begin{figure}
\centering
\includegraphics[width=0.95\textwidth]{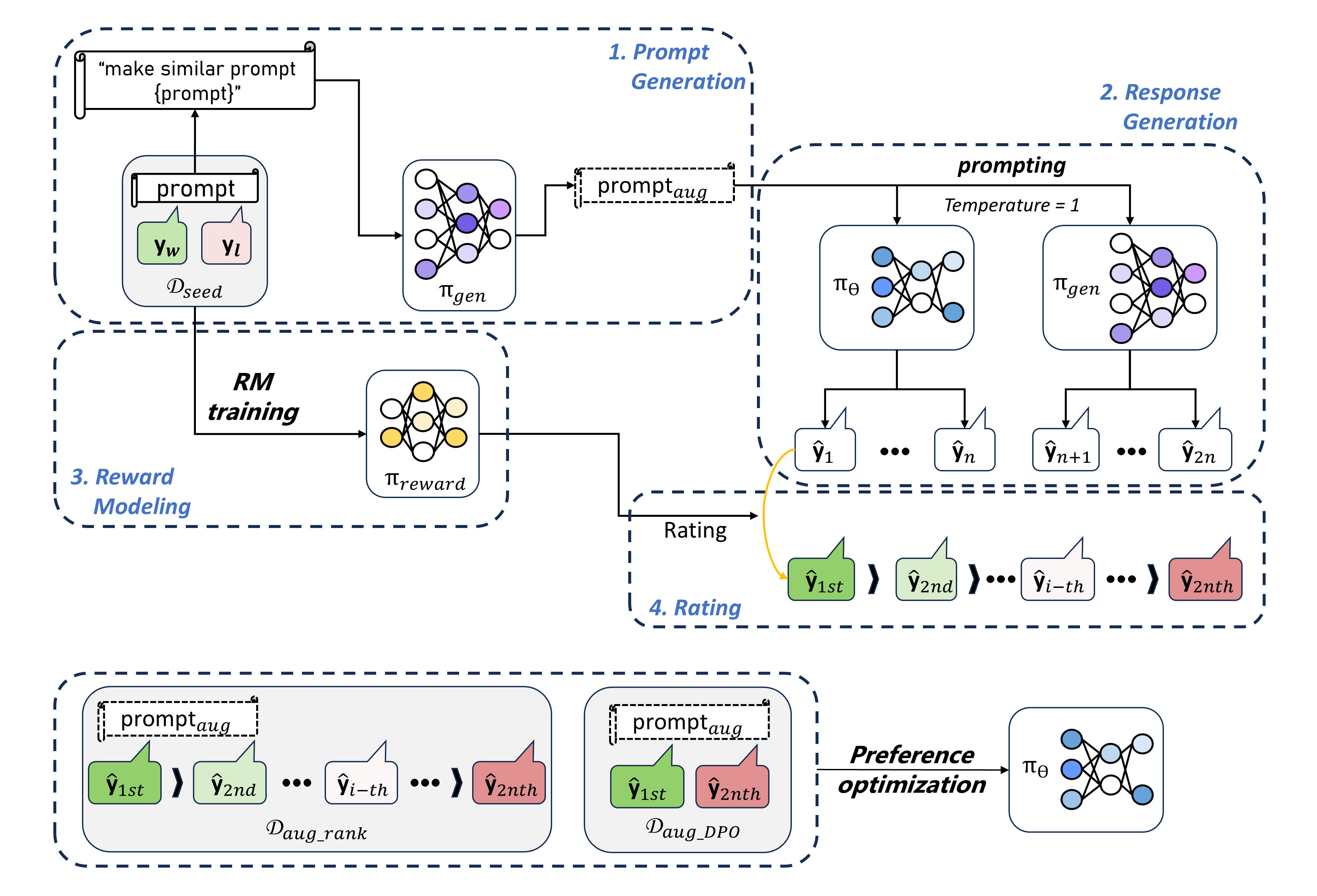}
\caption{The preference dataset augmentation is conducted through four stages. In each stage, all data is generated, trained, and evaluated by the model without human intervention.} \label{figure:process}
\end{figure}

Dataset augmentation is performed using four components: the seed preference dataset $\mathcal{D}_{seed} = \{x^{p}, y^{p}_{w},y^{p}_{l}\}^{P}_{p=1}$, the data generation model $\pi_{gen}$, the policy model $\pi_{\theta}$, and the reward model $\pi_{reward}$ for preference evaluation.

As the first step of data augmentation, we generate prompts from the seed data. A subset of the prompts is sampled and provided as examples to $\pi_{gen}$, which is then tasked with generating similar prompts $\hat{x} \sim \pi_{gen}(y|X, x_{seed})$ through few-shot learning. Here, \textit{X} represents prompts used for data generation (Appendix \ref{appendix:prompt_augmentation}). The quality of the generated $\hat{x}$ depends on the performance of $\pi_{gen}$, so a high-performing model is chosen for $\pi_{gen}$.

In the second step, responses $\hat{y}$ are generated for the prompts $\hat{x}$. Both $\pi_{gen}$ and $\pi_{\theta}$ are utilized, and each model generates \textit{n} responses $\hat{y}$. Here, $\hat{x}$ represents the generated prompt, and $\hat{y}$ represents the generated response.

\begin{center}
$\{ \hat{y}_{1}, \hat{y}_{2}, ..., \hat{y}_{n}\}\sim\pi_{gen}(y|\hat{x})$
\end{center}
\begin{center}
$\{\hat{y}_{n+1}, \hat{y}_{n+2}, ..., \hat{y}_{2n}\} \sim \pi_{\theta}(y|\hat{x})$
\end{center}

In the third step, the reward model $\pi_{reward}$ is trained using the seed preference dataset. The reward model evaluates (x, y) pairs and outputs a scalar value representing how much the response is preferred for the given prompt. Using the trained reward model, the preferences for the responses $\hat{y}$ are calculated. Based on the calculated preferences, a ranked preference dataset $\mathcal{D}_{rank} = \{\hat{x}^{p}, \hat{y}^{p}_{1}, \hat{y}^{p}_{2}, \cdot\cdot\cdot, \hat{y}^{p}_{2n}\}^{P}_{p=1}$ is created, where $\mathcal{D}_{rank}$ contains a total of $P$ prompts, each associated with \textit{2n} ranked responses. Fig \ref{figure:process} illustrates the overall augmentation process.

\subsection{Multi-DPO}
DPO is a simple yet highly effective optimization algorithm that ensures robust performance. However, DPO is limited in the amount of information it learns, as it operates with only two responses per prompt. While human preferences can be expressed in a binary manner, such as "preferred" and "dispreferred," they can also be represented in greater detail through the ranking of multiple responses.
One potential approach is to adapt the existing DPO algorithm to train the policy model $\pi_{\theta}$ using ranked response information. In this case, the ranked dataset can be transformed by taking adjacent pairs of responses: the higher-ranked response is labeled as "preferred," and the lower-ranked response as "dispreferred." This transformation enables the application of DPO to ranked datasets.
For example, a ranked dataset with responses ranked from 1st to \textit{N}-th place can be converted into \textit{N-1} preferred-dispreferred pairs per prompt, allowing DPO to be applied to the transformed dataset effectively.
\begin{equation}
    \big[ \max_{\pi_{\theta}}\mathbb{E}_{(x,y_{i},y_{i+1}) \sim \mathcal{D}_{rank}}[\log\sigma(\beta \frac{\pi_{\theta}(y_{i}|x)}{\pi_{ref}(y_{i}|x)} - \beta \frac{\pi_{\theta}(y_{i+1}|x)}{\pi_{ref}(y_{i+1}|x)})]\big]_{i=1}^{\textit{N}-1} 
\end{equation}
However, this approach requires a significant increase in training effort. A ranked preference dataset $\mathcal{D}_{rank} = \{x^{p}, y^{p}_{1}, y^{p}_{2}, \cdot\cdot\cdot, y^{p}_{N}\}^{P}_{p=1}$, consisting of \textit{P} prompts, each with \textit{N} responses, would be converted into a DPO dataset $\mathcal{D}_{DPO} = \{x^{p}, y^{p}_{w}, y^{p}_{l}\}^{P \times (N-1)}_{p=1}$ containing $\textit{P} \times \textit{(N-1)}$ pairs. This transformation increases the amount of training data by a factor of (\textit{N-1}), significantly raising the computational and training requirements.

Therefore, we propose Multi-DPO, which enables the model to learn rank information simultaneously. Multi-DPO begins by redefining the objective of equation \ref{equation:RLHF_BT}. The $r^{*}$ in equation \ref{equation:RLHF_BT} can be replaced with the learnable $r(x,y)$ from equation \ref{equation:new_reward}.

\begin{equation}
    p^{*}(y_{1}>y_{2}|x) = \frac{1}{1+\exp(r(x,y_2) - r(x,y_{1}))}    
\end{equation}
$p^{}(y_{1}>y_{2}>\cdot\cdot\cdot > y_{n}|x)$ can be expressed as the product of $p^{}$ values for all adjacent $y$ pairs in the ranking.
\begin{align*}
    p^{*}(y_{1}>y_{2}>\cdot\cdot\cdot > y_{n}|x) &= \prod_{i=1}^{n-1} p^{*}(y_{i}>y_{i+1}|x)
    \\&=\prod_{i=1}^{n-1}(\frac{1}{1+\exp(r(x,y_{n+1}) - r(x,y_{n}))})
\label{equation:multiDPO1}
\end{align*}
Here, the objective of $r(x, y)$ is to maximize $p^{*}(y_{1}>y_{2}>\cdot\cdot\cdot > y_{n}|x)$, and the optimal $r(x, y)$ is defined as follows.
\begin{subequations}\label{equation:multi-DPO2}
\begin{align*}
    r^{*}(x,y) &=\operatorname*{argmin}_r (\prod_{i=1}^{n-1}(1+\exp(r(x,y_{n+1}) - r(x,y_{n}))) \tag{\ref{equation:multi-DPO2}}
\end{align*}    
\end{subequations}

Equation \ref{equation:multi-DPO2} can be approximated as follows. Further details on this approximation are provided in Appendix \ref{appendixA}.
\begin{subequations}\label{equation:multi-DPO3}
\begin{align*}
r^{*}(x,y) &\approx \operatorname*{argmin}_r (\prod_{i=1}^{n-1}(\exp(r(x,y_{n+1}) - r(x,y_{n}))))\\
        &=\operatorname*{argmin}_r(\exp(r(x,y_{n}) - r(x,y_{1})))\\
        &=\operatorname*{argmin}_r(r(x,y_{n}) - r(x,y_{1}))\tag{\ref{equation:multi-DPO3}}
\end{align*}
\end{subequations}

Equation \ref{equation:multi-DPO3} shares the same objective as standard DPO. Through equation \ref{equation:multi-DPO3}, we can observe that when training DPO with a dataset $\mathcal{D} = {x, y_{w}, y_{l}}$, it approximately incorporates the information between $y_{w}$ and $y_{l}$. This implies that the greater the difference between $y_{w}$ and $y_{l}$, the more information can be learned. This supports the importance of the performance of the data generation model $\pi_{gen}$ in our study.

However, equation \ref{equation:multi-DPO3} does not capture the detailed information of responses other than $y_{1}$ and $y_{n}$. To address this, we constructed the objective function using all possible pairwise comparisons among the responses, beyond just comparing adjacent responses in the ranking. All possible pairwise comparisons among the responses can be expressed as follows.
\begin{align}
    &p^{*}(y_{1}>y_{2}|x) \label{equation:multi-DPO4_1}
    \\
    &p^{*}(y_{1}>y_{3}|x)   \quad p^{*}(y_{2}>y_{3}|x) \label{equation:multi-DPO4_2}
    \\  \nonumber
    &\qquad \cdot\cdot\cdot  \quad\quad\quad\quad\qquad  \cdot\cdot\cdot\\
    &p^{*}(y_{1}>y_{n}|x) \quad  p^{*}(y_{2}>y_{n}|x) \quad\cdot\cdot\cdot \quad p^{*}(y_{n-1}>y_{n}|x) \nonumber
\end{align}

Through all these comparisons, the optimal reward model can be expressed as follows(A total of \textit{(n-1)}! comparisons are conducted.).
\begin{equation}
    r^{*}(x,y) = \operatorname*{argmax}_r \prod_{i=1}^{n-1}\prod_{k=i}^{n}(\exp(r(x,y_{i}) - r(x,y_{k})))
\end{equation}
As proven in equation \eqref{equation:multi-DPO3}, when the same response $\hat{y}_{k}$ appears across multiple terms, the preferred $\hat{y}_{k}$ and dispreferred $\hat{y}_{k}$ cancel each other out. If $\hat{y}_{k}$ appears repeatedly as preferred or repeatedly as dispreferred, their contributions are aggregated. For example, $( >y_{2})$ in equation \eqref{equation:multi-DPO4_1} can cancel with $(y_{2}>)$ in equation \eqref{equation:multi-DPO4_2}. Assuming there are $n$ responses, the rule for canceling responses is as follows.

($y_{1}>$) is not canceled and is aggregated \textit{n-1} times: \textit{n-1}\newline

($y_{2}>$) is canceled once and aggregated \textit{n-2} times: \textit{n-2-1}\newline

($y_{i}>$) is canceled \textit{i-1} times and aggregated \textit{n-i} times: \textit{(n-i) - (i-1)}
\newline
\newline
Based on this rule, $r^{*}(x,y)$ is defined as follows.
\begin{equation}
    r^{*}(x,y) = \operatorname*{argmax}_r\sum\limits_{i=1}^{n}(n-2i+1)(r(x,y_{i}))
\end{equation}

Finally, our policy objective can be expressed in its normalized form as follows.
\begin{equation}
    \mathcal{L}_{multiDPO} = \max_{\pi_{\theta}}\mathbb{E}_{(x,y_{1},y_{2}, ...,y_{n}) \sim \mathcal{D}_{rank}}[\log\sigma(\beta\sum\limits_{i=1}^{n}\frac{(n-2i+1)}{(n-1)} \frac{\pi_{\theta}(y_{i}|x)}{\pi_{ref}(y_{i}|x)})]
\label{equation:MDPO_tuning}
\end{equation}
When \textit{n=2}, this objective aligns with the objective of DPO.

\section{Experiment}
\subsection{Datasets}
In this experiment, we used \textit{orca dpo pairs} \cite{orca_dpo} \cite{mukherjee2023orcaprogressivelearningcomplex}, which consists of approximately 13,000 {prompt, chosen response, rejected response} pairs. This dataset includes instructions related to translation, writing, common sense, and mathematical reasoning, summary but does not include instructions related to coding. We used this dataset as the seed dataset for data augmentation and also for DPO training.
\subsection{Augmentation}
Dataset augmentation is divided into four stages: prompt augmentation, response augmentation, reward modeling and rating . In the prompt augmentation stage, prompts from the seed dataset are provided as examples to the generation model $\pi_{gen}$, which generates similar prompts $\hat{x}$. For each seed prompt, three augmented prompts were generated, resulting in a total of 39,000 new prompts. Among these, 2,000 prompts that were excessively short were excluded, leaving a final set of 37,000 prompts. The generation model $\pi_{gen}$ used for this process was GPT-3.5-turbo.

The generated $\hat{x}$ is used for response augmentation. Each $\hat{x}$ is prompted to $\pi_{gen}$ and $\pi_{\theta}$ to generate responses $\hat{y}$. For each prompt, two responses were generated per model ($\pi_{gen}$ and $\pi_{\theta}$), resulting in a total of four responses per prompt. To ensure diversity in the responses, the temperature of each model was set to 1.

In the rating stage, rewards are assigned to $(\hat{x}, \hat{y})$ pairs using a reward model. The reward model can either be trained using $\mathcal{D}_{seed}$ or be a pretrained model. In our study, we used the pretrained \textit{RM-Gemma-7B}\cite{dong2023raft} as the reward model. The reward model rated the prompt and its four responses, and the rated responses were sorted based on their rewards. As a result, we obtained the ranked dataset $\mathcal{D}_{rank} = \{\hat{x}, \hat{y}_{1}, \hat{y}_{2}, \hat{y}_{3}, \hat{y}_{4}\}$.

\subsection{Training}
The policy model $\pi_{\theta}$ used in this study was the Mistral-SFT 7B model\cite{jiang2023mistral}. The model underwent 4-bit quantization \cite{han2015deep}, and DPO training was performed using LoRA\cite{hu2021loralowrankadaptationlarge} with a rank of 64 and alpha of 64. Training was conducted with a batch size of 2 for a total of 3 epochs, using a learning rate of 1e-6. 
We associated 4 responses with each prompt, and as a result, the final loss is as follows, in accordance with equation \eqref{equation:MDPO_tuning}.
\begin{align}
    \mathcal{L}_{mDPO} =  \nonumber
    \\\max_{\pi_{\theta}}\mathbb{E}_{(x,y_{1},y_{2}, y_{3}, y_{4}) \sim \mathcal{D}_{rank}^{n=4}}[
    &\log\sigma(\beta\frac{\pi_{\theta}(y_{1}|x)}{\pi_{ref}(y_{1}|x)} + 0.33\beta\frac{\pi_{\theta}(y_{2}|x)}{\pi_{ref}(y_{2}|x)} \nonumber \\ 
    &- 0.33\beta\frac{\pi_{\theta}(y_{3}|x)}{\pi_{ref}(y_{3}|x)}-\beta\frac{\pi_{\theta}(y_{4}|x)}{\pi_{ref}(y_{4}|x)})]
\label{equation:MDPO_tuning_4response}
\end{align}

We applied both general DPO and Multi-DPO to three datasets $\mathcal{D}_{seed}$, $\mathcal{D}_{aug}$, $\mathcal{D}_{seed+aug}$ resulting in a total of six models. By comparing these six models, we evaluated the effectiveness of dataset augmentation and Multi-DPO(MDPO).

\section{Results}

We evaluated the models based on their instruction-following capabilities. During the evaluation, the models were provided with instructions and tasked to perform accordingly. The results of their execution were assessed by an LLM. This evaluation of instruction-following ability was divided into two categories: single-turn evaluation and multi-turn evaluation.

\subsection{Model}

\begin{table}[]
\caption{The format of the datasets and the size of the datasets used for training each model.}\label{table:model_dataset}
\centering
\begin{tabular}{c|c|c|c|c|}
\cline{2-5}
                                     & Dataset & N datasets &train steps & train time \\ \hline
\multicolumn{1}{|c|}{$DPO_{seed}$}      & $ (x, y_{w}, y_{l})$       & 13000 &2400 & 10h 17m \\ \hline
\multicolumn{1}{|c|}{$MDPO_{seed}$}     & $ (x, y_{w}, y_{l},{y}_{1},{y}_{2})$     & 13000 & 2400 & 18h 42m \\ \hline
\multicolumn{1}{|c|}{$DPO_{aug}$}       & $ (\hat{x}, \hat{y}_{1}, \hat{y}_{4})$   &37000    & 6500 & 23h 12m \\ \hline
\multicolumn{1}{|c|}{$MDPO_{aug}$}      & $ (\hat{x},\hat{y}_{1}, \hat{y}_{2},\hat{y}_{3},\hat{y}_{4})$   & 37000   & 6500 &  38h 23m \\ \hline
\multicolumn{1}{|c|}{$DPO_{seed+aug}$}  & $ (\mathcal{D}_{DPO_{seed}}+\mathcal{D}_{DPO_{aug}})$    & 50000  & 9000 & 31h 53m\\ \hline
\multicolumn{1}{|c|}{$MDPO_{seed+aug}$} & $ (\mathcal{D}_{MDPO_{seed}}+\mathcal{D}_{MDPO_{aug}})$   & 50000   & 9000 & 61h 27m  \\ \hline
\end{tabular}
\end{table}

We named the models by combining the dataset used and the training method. For instance, $DPO_{seed}$ refers to a model trained on the seed dataset using DPO. Table~\ref{table:model_dataset} shows the format of datasets used for each model. The dataset for $MDPO_{seed}$ only applies response augmentation (${y}{1}, {y}{2}$) to the seed prompt, with no augmentation applied to the prompt itself. According to Table~\ref{table:model_dataset}, when using the same prompt, \textit{MDPO} took approximately 1.6 to 2 times longer to train compared to DPO. When training all ranks informatino with DPO, the data size increases by a factor of 6((n-1)!), and the training time increases proportionally. Therefore, it can be concluded that learning rank information with \textit{MDPO} is more efficient than training with DPO. The training was conducted using a single A6000 GPU. 
\subsection{Single-turn evaluation}
\begin{table}[]
\caption{The results of the single-turn test, AlpacaEval. AlpacaEval evaluates models based on the win rate against the base model.}\label{table:AlpacaEVAL}
\centering
\begin{tabular}{c|r|r|}
\cline{2-3}
                                     & \multicolumn{1}{l|}{\textit{\textbf{AlpacaEval 2.0}}} & \multicolumn{1}{l|}{\textit{\textbf{AlpacaEval 1.0}}} \\ \hline
\multicolumn{1}{|c|}{$DPO_{seed}$}      & 4.31                                                  & 41.22                                                 \\ \hline
\multicolumn{1}{|c|}{$MDPO_{seed}$}     & 8.40                                                   & 69.32                                                 \\ \hline
\multicolumn{1}{|c|}{$DPO_{aug}$}       & 12.96                                                 & 79.23                                                 \\ \hline
\multicolumn{1}{|c|}{$MDPO_{aug}$}      & 14.49                         & 79.85                                                 \\ \hline
\multicolumn{1}{|c|}{$DPO_{seed+aug}$}  & 12.36                                                 & 77.49                                                 \\ \hline
\multicolumn{1}{|c|}{$MDPO_{seed+aug}$} & 12.50                                              & 79.48                                                 \\ \hline
\end{tabular}
\label{table:alpaca_eval}
\end{table}
In the single-turn instruction-following evaluation, the models are assessed based on their performance in a single round of instruction and response. The AlpacaEval benchmark \cite{alpaca_eval}\cite{dubois2024lengthcontrolledalpacaevalsimpleway} dataset was used for this purpose. The AlpacaEval dataset consists of 805 instructions, a baseline model, and autoannotators. The evaluated model is prompted with the 805 instructions and generates responses. These responses are then compared with the outputs of the baseline model. Using the results from both the evaluated model and the baseline model, the autoannotators select the better response. The performance of the model is determined by its win rate against the baseline model. AlpacaEval is divided into two versions: AlpacaEval 1.0 and AlpacaEval 2.0. In AlpacaEval 1.0, the baseline model is text-davinci-003, and the autoannotators are GPT-4-turbo\cite{openai2024gpt4technicalreport}. In AlpacaEval 2.0, both the baseline model and the autoannotators are GPT-4-turbo.

As shown in Table~\ref{table:AlpacaEVAL}, there are significant performance differences between models in the AlpacaEval benchmark. First, the two models trained on $\mathcal{D}_{seed}$, namely $DPO_{seed}$ and $MDPO_{seed}$, achieved win rates of 4.31\%/41.22\% and 8.40\%/69.32\%, respectively, in AlpacaEval 1.0 and 2.0. These results indicate that optimizing with multiple responses significantly improves performance compared to optimization with only two responses.
A similar trend is observed for models trained on the generated dataset $\mathcal{D}_{aug}$. While $DPO_{aug}$ achieved win rates of 12.96\%/79.23\%, $MDPO_{aug}$ achieved slightly better results with 14.49\%/79.85\%. These findings further confirm the superiority of \textit{MDPO} over DPO.

While the differences between \textit{DPO} and \textit{MDPO} are significant, the impact of the dataset is even more pronounced. When using \textit{DPO}, the win rate increased substantially from 4.31\%/41.22\% with $DPO_{seed}$ to 12.96\%/79.23\% with $DPO_{aug}$. Similarly, for \textit{MDPO}, the win rate rose significantly from 8.40\%/69.32\% with $MDPO_{seed}$ to 14.49\%/79.85\% with $MDPO_{aug}$. These results demonstrate the effectiveness of our dataset generation framework. Furthermore, they indicate that improvements driven by the dataset have a greater impact on performance than those achieved through advancements in training methods.In summary, the findings suggest that training with \textit{MDPO} on the generated dataset yields the best performance.

When combining the two datasets, the results deviated from our expectations. We initially hypothesized that using $DPO_{seed+aug}$ would yield the highest performance. However, the actual results showed that $DPO_{seed+aug}$ outperformed $DPO_{seed}$ but fell slightly short of $DPO_{aug}$. Several potential reasons could explain this outcome.
First, the lower quality of $DPO_{seed}$ may have contributed to the results. If $DPO_{seed}$ underperforms, it could counteract the improvements achieved by $DPO_{aug}$, thereby reducing the overall performance. The suboptimal performance of $DPO_{seed}$ supports this hypothesis. Second, the issue of self-enhancement bias could be a factor. According to Zheng et al.\cite{zheng2023judging}, LLMs tend to favor responses they generate themselves. In our setup, half of the augmented data was generated by GPT-3.5-turbo, and the evaluation was conducted by GPT-4-turbo. Consequently, models trained exclusively on responses generated by GPT-3.5-turbo, which shares similarities with the evaluator, may have experienced a self-enhancement bias, resulting in higher performance than models trained on combined datasets.

\subsection{Multi-turn evaluation}
\begin{table}[]
\caption{Results of the multi-turn test, MT-bench. The models respond to questions across 8 categories in a multi-turn manner and are evaluated on a scale from 0 to 10.}\label{table:MT_bench}
\centering
\begin{tabular}{l|ccccccccc}
\multicolumn{1}{r|}{\textbf{category}} & \textbf{Cod}  & \textbf{Ext}  & \textbf{Hum}  & \textbf{Math} & \textbf{Rsn}  & \textbf{Rol}  & \textbf{STEM} & \textbf{Wrt}                       & \textbf{mean} \\
\multicolumn{1}{r|}{\textbf{model}}    &               &               &               &               &               &               &               &                                    &               \\ \hline
$DPO_{seed}$                           & \textbf{4.62} & 5.75          & 9.32          & 2.02          & 5.03          & 6.74          & 7.64          & \multicolumn{1}{c|}{7.58}          & 6.09          \\
$MDPO_{seed}$                          & 4.30          & 7.12          & 9.48          & 3.38          & 5.15          & 8.00          & 8.12          & \multicolumn{1}{c|}{8.50}          & 6.76          \\
$DPO_{aug}$                            & 3.88          & \textbf{7.92} & \textbf{9.80} & 3.50          & 5.42          & \textbf{8.14} & 8.66          & \multicolumn{1}{c|}{9.02}          & \textbf{7.04} \\
$MDPO_{aug}$                           & 3.65          & 7.15          & \textbf{9.80} & \textbf{3.90} & 5.20          & 8.05          & \textbf{9.30} & \multicolumn{1}{c|}{8.60}          & 6.96          \\
$DPO_{seed+aug}$                       & 3.65          & 7.10          & 9.40          & 3.65          & 5.80          & 8.05          & 8.18          & \multicolumn{1}{c|}{8.85}          & 6.83          \\
$MDPO_{seed+aug}$                      & 4.08          & 7.52          & 9.62          & 3.45          & \textbf{6.25} & 7.75          & 8.26          & \multicolumn{1}{c|}{\textbf{9.05}} & 7.00         
\end{tabular}
\end{table}
Multi-turn instruction-following evaluation assesses models over multiple stages of instruction and response. MT-bench\cite{zheng2023judging} is used for this purpose. MT-bench consists of 80 high-quality multi-turn questions, and the responses to these questions are evaluated by an LLM, such as GPT-4. The questions are categorized into eight domains (Writing, Roleplay, Reasoning, Math, Coding, Extraction, STEM, and Humanities), and the judge LLM assigns a score between 0 and 10 for each response within these domains.

Table~\ref{table:MT_bench} presents the results for the MT-bench evaluation. Except for the performance of $DPO_{aug}$, the results follow a pattern similar to the single-turn test. Overall, \textit{MDPO} outperformed \textit{DPO}, and models trained on $\mathcal{D}_{aug}$ generally exhibited superior performance. These findings further validate the effectiveness of the proposed model and approach.

The results in Table \ref{table:MT_bench} also illustrate how the characteristics of the dataset influence performance across different categories. The \textit{orca dpo pair} includes categories such as writing, common sense, reasoning, and mathematics but does not cover coding. In categories included in dataset, performance improved with training. However, in the coding category, which was not represented in the dataset, performance declined as training progressed. Considering these dataset characteristics, while $DPO_{aug}$ achieved the highest overall scores, \textit{MDPO} models generally performed better in the primary domains of \textit{orca dpo pair}.

\section{Conclusion}
In this study, we explored methods for generating and training datasets to optimize LLMs. The datasets were generated based on a seed dataset and rated using a reward model. The proposed optimization technique is based on multi-response learning, which simplifies the potentially complex process of multi-response training through cancellation mechanisms.
Consequently, the contributions of this study can be divided into two main aspects: dataset augmentation and training methodology.

The experiments were divided into two parts: validation of the augmented dataset and validation of the proposed training method. The results demonstrated that the dataset we generated significantly improved model performance, indicating the effectiveness of the dataset generation process proposed in this study.
Additionally, the proposed multi-DPO generally outperformed traditional DPO under the same dataset conditions. This performance gap was particularly pronounced when the dataset size was small. These findings suggest that training with only two responses limits the amount of learnable data, and multi-DPO effectively addresses this limitation.

While our study demonstrated superior performance compared to previous approaches, it has several limitations. One of the key advantages of DPO is its ability to train without requiring a reward model; however, our method reintroduces the need for a reward model. Additionally, the proposed objective function converges to an approximate solution rather than an optimal one. As part of ongoing research, we are exploring simpler and more optimized approaches to address these limitations and anticipate finding improved solutions in the near future.


\appendix
\section{derivation of $p^{*}(y_{1}>y_{2}>\cdot\cdot\cdot > y_{n} | x)$.}
\label{appendixA}

To derive $p^{*}(y_{1}>y_{2}>\cdot\cdot\cdot > y_{n} | x)$, we first consider the case where there are three responses (n=3). Assuming that only adjacent responses in the ranking are compared:

\begin{align}
    p^{*}(y_{1}>y_{2}>y_{3} | x) &= p^{*}(y_{1}>y_{2} | x)p^{*}(y_{2}>y_{3} | x)\nonumber \\
    &=(\frac{\exp(r^{*}(x,y_{1}))}{\exp(r^{*}(x,y_{1})) + \exp(r^{*}(x,y_{2}))})(\frac{\exp(r^{*}(x,y_{2}))}{\exp(r^{*}(x,y_{2})) + \exp(r^{*}(x,y_{3}))}) \nonumber\\
    &=(\frac{1}{1 + \exp(r^{*}(x,y_{2}) -r^{*}(x,y_{1}))})(\frac{1}{1 + \exp(r^{*}(x,y_{3}) -r^{*}(x,y_{2}))}) \label{equation:APPENDIX_1}
\end{align}

Maximizing equation \eqref{equation:APPENDIX_1} can be interpreted as minimizing the reciprocal of the function.

\begin{align}
    &maximize(\frac{1}{1 + \exp(r^{*}(x,y_{2}) -r^{*}(x,y_{1}))})(\frac{1}{1 + \exp(r^{*}(x,y_{3}) -r^{*}(x,y_{2}))}) \nonumber \\
    &=minimize(1 + \exp(r^{*}(x,y_{2}) -r^{*}(x,y_{1})))(1 + \exp(r^{*}(x,y_{3}) -r^{*}(x,y_{2}))) \label{equation:APPENDIX_2}
\end{align}

Here, substituting $A=\exp(r^{}(x,y_{1}))$, $B=\exp(r^{}(x,y_{2}))$, and $C=\exp(r^{*}(x,y_{3}))$, equation \eqref{equation:APPENDIX_2} can be expressed as follows.

\begin{align}
    &(1 + \exp(r^{*}(x,y_{2}) -r^{*}(x,y_{1})))(1 + \exp(r^{*}(x,y_{3}) -r^{*}(x,y_{2}))) \nonumber \\
    &= (1 + \frac{B}{A})(1 + \frac{C}{B}) \label{equation:APPENDIX_3}  \\
    &= (1+\frac{B}{A}+\frac{C}{B}+\frac{C}{A}) 
\end{align}

\begin{lemma}
\label{lemma1}
    $minimize(\frac{C}{A}) \approx minimize(1+\frac{B}{A}+\frac{C}{B}+\frac{C}{A})$
\begin{proof}
    In $(1+\frac{B}{A}+\frac{C}{B}+\frac{C}{A})$, the term 1 is a constant, and $\frac{C}{A}$ is the term to be minimized. Therefore, the proof focuses on the relationship between $\frac{C}{A}$ and $(\frac{B}{A}+\frac{C}{B})$.
    
    Since $A, B, C$ are exponential functions, they are positive real numbers. Consequently, $\frac{B}{A}$ and $\frac{C}{B}$ are also positive real numbers. Therefore, the following inequality holds.
    
    \begin{equation}
        \frac{B}{A}+\frac{C}{B} \geq 2\sqrt{\frac{B}{A}\frac{C}{B}} = 2\sqrt{\frac{C}{A}}
    \end{equation}
    
    Furthermore, if $A\geq B\geq C$ is assumed, $\frac{B}{A}+\frac{C}{B}$ reaches its maximum value when $B=A$ or $B=C$.

    \begin{equation}
        \frac{B}{A}+\frac{C}{B} \leq 1+\frac{C}{A}
    \end{equation}

    In summary, $\frac{B}{A}+\frac{C}{B}$ can be expressed as follows.
    \begin{equation}
        2\sqrt{\frac{C}{A}} \leq \frac{B}{A}+\frac{C}{B} \leq 1+\frac{C}{A}
    \label{equation:APPENDIX_4}
    \end{equation}
    According to equation \eqref{equation:APPENDIX_4}, $\frac{C}{A}$ determines the lower and upper bounds of $\frac{B}{A}+\frac{C}{B}$. Thus, by minimizing $\frac{B}{A}$, it is possible to approximately minimize $\frac{B}{A}+\frac{C}{B}$. Additionally, assuming $B$ is fixed, minimizing $\frac{C}{A}$ also minimizes $\frac{B}{A}+\frac{C}{B}$.    
\end{proof}
\end{lemma}

By lemma \ref{lemma1}, the problem of minimizing equation \eqref{equation:APPENDIX_3} can be expressed as follows.

\begin{equation}
\label{equation:APPENDIX_5}
    minimize(1 + \frac{B}{A})(1 + \frac{C}{B}) \approx minimize(\frac{B}{A})(\frac{C}{B})
\end{equation}

Applying equation \eqref{equation:APPENDIX_5} to compare all adjacent responses from $y_{1}$ to $y_{n}$ yields the following.

\begin{align}
\label{equation:APPENDIX_6}
minimize (\prod_{i=i}^{n-1}(1+\exp(r(x,y_{i+1}) - r(x,y_{i})))&\approx minimize(\prod_{i=i}^{n-1}\exp(r(x, y_{n+1})-r(x, y_{n}))) \nonumber \\
&=minimize((\exp(r(x, y_{n})-r(x, y_{1}))))
\end{align}

Applying equation \eqref{equation:APPENDIX_6} to compare all comparable responses from $y_{1}$ to $y_{n}$ can be expressed as follows.
\begin{equation}
    minimize (\prod_{i=1}^{n-1}\prod_{k=i}^{n}(\exp(r(x,y_{i}) - r(x,y_{k})))) \approx minimize(\exp(\sum\limits_{i=1}^{n} (n-2i+1)r(x,y_{i})))
\end{equation}

\section{MT-bench}
\begin{figure}
\includegraphics[width=0.9\textwidth]{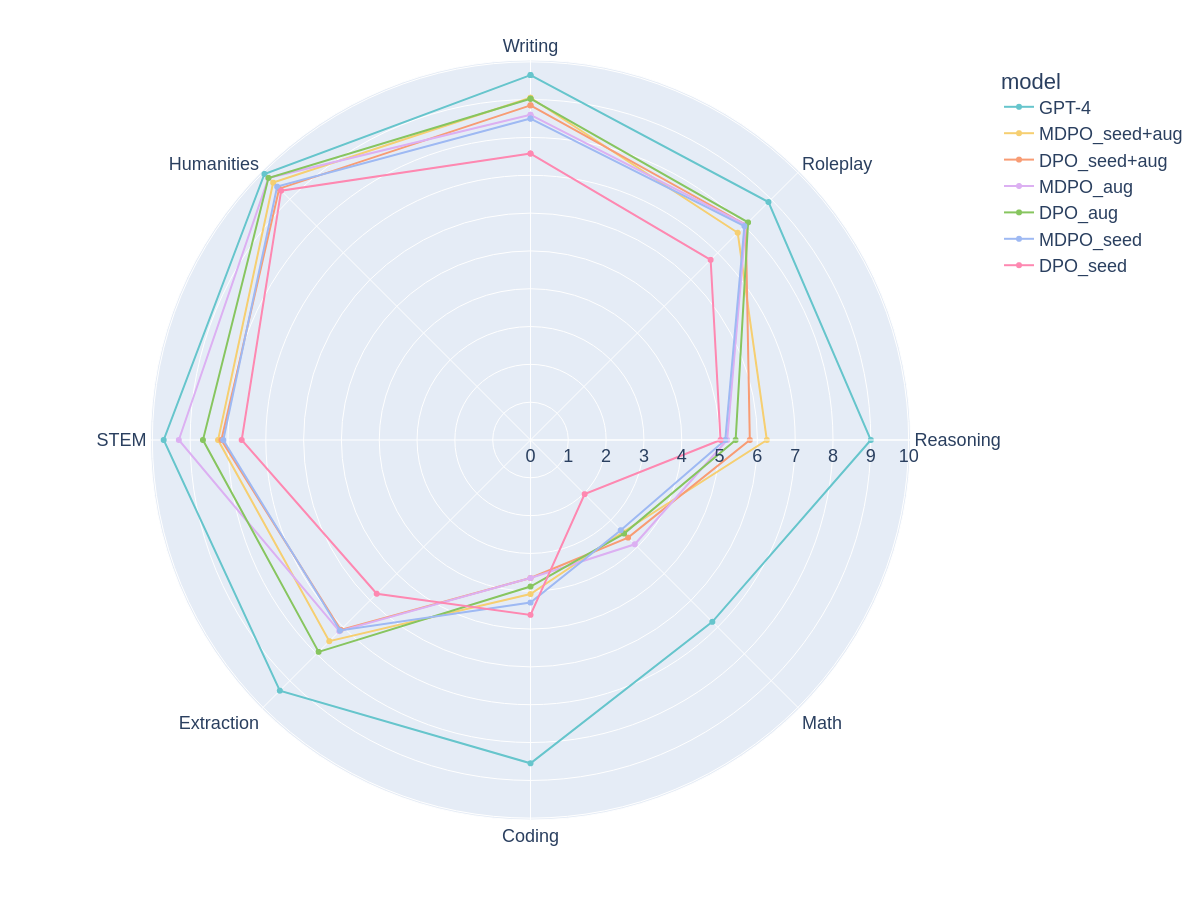}
\caption{Results of MT-bench test.} \label{figure:MT-bench}
\end{figure}
Figure \ref{figure:MT-bench} shows the results of the multi-turn test, the MT-bench test. Each model exhibited strengths in different areas. While most models scored higher than $DPO_{seed}$ in various domains, they recorded lower scores in the Coding domain. This phenomenon seems to be due to the absence of Coding-related problems in the training data, \textit{orca dpo pair}, leading to a loss of performance in the Coding-related abilities.

\section{prompt augmentation}
\label{appendix:prompt_augmentation}

\begin{table}[]
\begin{tabular}{|c|l|}
\hline
\textbf{model}   & gpt-3.5-turbo                                                                                                                                                                                                                                     \\ \hline
\textbf{role}    & user                                                                                                                                                                                                                                                                                                                                                                                                                                                                                                                       \\ \hline
\textbf{content} & \begin{tabular}[c]{@{}l@{}}'''Create a new instructions by referring to the following sample instructions.\\ Never reveal the answer and only provide the instructions.\\ The format should be similar to "sample instruction," \\
but the content must be entirely independent.\\ 
sample instruction1 : "\{\}"\\ 
Using the above instruction as a reference, \\
create a completely independent new instruction.\\ The response to the instruction should never be included.\\ new instruction: '''.format(sample)\end{tabular} \\ \hline
\end{tabular}
\caption{A prompt to generate new prompts from a \textit{sample}}
\label{table:prompt_augment}
\end{table}
Table \ref{table:prompt_augment} shows the method of generating new prompts using a generative model (gpt-3.5-turbo). The generative model takes one seed prompt as input at a time and generates a new prompt. We applied this process three times to each seed prompt, effectively augmenting the seed prompts by a factor of three. Subsequently, 2,000 prompts that were too short were removed.
\end{document}